\documentclass[lettersize,journal]{IEEEtran}
\usepackage{amsmath,amsfonts}
\usepackage{algorithmic}
\usepackage{algorithm}
\usepackage{array}
\usepackage[caption=false,font=normalsize,labelfont=sf,textfont=sf]{subfig}
\usepackage{textcomp}
\usepackage{stfloats}
\usepackage{url}
\usepackage{verbatim}
\usepackage{graphicx}
\usepackage{cite}
\begin{document}

\title{Transforming Vehicle Diagnostics: A Multimodal Approach to Error Patterns Prediction}

\author{Hugo Math$^{1}$, Rainer Lienhart$^{2}$

\thanks{$^{1}$Hugo Math is with Augsburg University, Augsburg 86159, Germany.}
\thanks{$^{2}$Prof. Dr. Rainer Lienhart is with Augsburg University, Augsburg 86159, Germany. Head of Chair for Machine Learning \& Computer Vision.}
}


\maketitle

\begin{abstract}
Accurately diagnosing and predicting vehicle malfunctions is crucial for maintenance and safety in the automotive industry. While modern diagnostic systems primarily rely on sequences of vehicular Diagnostic Trouble Codes (DTCs) registered in On-Board Diagnostic (OBD) systems, they often overlook valuable contextual information such as raw sensory data (e.g., temperature, humidity, and pressure). This contextual data, crucial for domain experts to classify vehicle failures, introduces unique challenges due to its complexity and the noisy nature of real-world data. This paper presents BiCarFormer: the first multimodal approach to multi-label sequence classification of error codes into error patterns that integrates DTC sequences and environmental conditions. BiCarFormer is a bidirectional Transformer model tailored for vehicle event sequences, employing embedding fusions and a co-attention mechanism to capture the relationships between diagnostic codes and environmental data. Experimental results on a challenging real-world automotive dataset with 22,137 error codes and 360 error patterns demonstrate that our approach significantly improves classification performance compared to models that rely solely on DTC sequences and traditional sequence models. This work highlights the importance of incorporating contextual environmental information for more accurate and robust vehicle diagnostics, hence reducing maintenance costs and enhancing automation processes in the automotive industry.
\end{abstract}


\section{Introduction}
\begin{figure}[!t]
      \centering
      \includegraphics[width=3.3in]{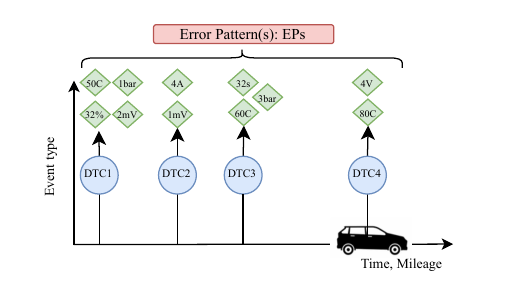}
      \caption{Error Pattern (EP) prediction based on past Diagnostic Trouble Codes (DTCs) and environmental conditions (e.g., temperature, voltage, \dots).}
      \label{fig:dtc_env_car}
\end{figure}

\IEEEPARstart{M}{odern} intelligent vehicles generate vast amounts of data. They are typically reported as events occurring irregularly over time. Some events happen simultaneously, while others are spaced unevenly. They often include redundant and noisy categorical and numerical features, highlighting the need to harness them effectively for improving vehicle diagnostics and predictive maintenance.
In this work, we focus on processing and analyzing multivariate, irregular event streams generated by modern vehicles — an area where more research is needed. These event sequences, known as Diagnostic Trouble Codes (DTCs), capture discrete events both in time and space during the life of a vehicle. DTCs are preferred over raw sensor data because they offer more structured, discrete information, making them easier to analyze.
However, when classifying these DTC sequences to identify critical failures such as error patterns (EPs), domain experts often rely on additional contextual data (Figure \ref{fig:dtc_env_car}) such as the environmental conditions of the vehicle (e.g., temperature, pressure, and voltage readings).
EPs are defined by domain experts after observing a good amount of DTC sequences. This makes them much more precise about the critical failure that the car is having. While some DTCs can also be noisy and
redundant events about recurring errors (e.g.,
software updates, electrical issues), EPs characterize a whole error sequence
consisting of precise vehicle failures (e.g., engine or battery
failures).
The current research relies solely on DTCs to infer either the next DTCs using Transformers \cite{attention, Hafeez2024DTCTranGruIT} and RNNs \cite{faultpredmultivariatevehicule} or the next EPs \cite{math2024harnessingeventsensorydata, math2025oneshot, math2025towards}. Thus, there is an extensive need to make use of this additional contextual data. 
While this secondary data can enhance the classification or clustering of EPs, it poses integration challenges due to its volume, variability, and high dimensionality.
Our goal is to uncover the appearance of Diagnostic Trouble Codes (DTCs) and environmental conditions that correlate with certain EPs. 
We demonstrate that our approach significantly improves the multi-label sequence classification of error patterns (EPs) in vehicles, outperforming traditional models that rely solely on DTCs \cite{Hafeez2024DTCTranGruIT, faultpredmultivariatevehicule} and classical ''sequence-to-sequence models'' such as BERT \cite{bert}.
By enhancing the prediction performance of EPs in vehicles, our approach enables the reduction of costly domain experts and transforms traditional vehicle diagnostics. We further explain classification predictions by interpreting cross-attention scores and show that BiCarFormer learns to detect fluctuations of continuous values to perform prediction, although encapsulating different units and thousands of different environmental conditions.
We emphasize the implications of real-world data to the proposed method and its potential further applications.
\textbf{Contributions.}
In this work, we introduce BiCarFormer, the first multimodal bidirectional Transformer designed to classify error patterns (EPs) in vehicles by integrating both Diagnostic Trouble Codes (DTCs) and environmental conditions. Our key contributions are:
\begin{itemize}
    \item Multimodal learning for EPs prediction: Unlike prior methods that rely solely on DTC sequences, we leverage multimodal sensor fusion, incorporating environmental conditions (e.g., temperature, voltage, pressure) to enhance classification accuracy.
    \item Co-Attention and event embedding fusion: We propose a novel fusion mechanism that efficiently integrates diverse sensory inputs, capturing meaningful interactions between DTCs and environmental data.
    \item Enhanced explainability: Through cross-attention score interpretation, we provide insights into how BiCarFormer detects fluctuations in continuous values, demonstrating its applicability to real-world vehicle diagnostics.
\end{itemize}
\section{Background and Related Work}
\subsection{Failure Detection in Vehicles}
\noindent Capturing machine failures has historically been done by machine learning models. Traditional Predictive Maintenance (PdM) includes techniques based on correlation \cite{7998309}, Bayesian networks \cite{LANGSETH200792}, decision tree models \cite{logbasedpredictivemaitnenancetree}, and more recently neural state-space models \cite{HE2023109598} to predict machine failures or remaining useful life (RUL). These methods rely on historical data to identify patterns and predict potential failures. Bayesian networks can model the probabilistic dependencies among various components of the vehicle. Some papers explore Deep Neural Networks (DNNs) for PdM tasks. For example, \cite{Zhou2023} uses a combination of convolutional, fully connected, and transformer modules within one architecture for the classification of machine event logs.
In the vehicle event sequence world, some papers suggest using RNN, LSTM \cite{faultpredmultivariatevehicule}, and Transformers \cite{attention} to leverage DTC data by forming sequences of Vehicular Diagnosis Trouble Codes (DTCs) and predicting the next token. \cite{Hafeez2024DTCTranGruIT} introduces DTC-TranGRU that further improves next-DTC prediction by integrating a Transformer and a recurrent unit. Nevertheless, predicting only the next DTC has its limitations in terms of real-world application and remains a difficult task due to the enormous amount of DTCs, closely matching a language as pointed out by \cite{math2024harnessingeventsensorydata}. Indeed, \cite{math2024harnessingeventsensorydata} suggests using the warranty data of vehicles to create a supervised learning problem where they attach error patterns (e.g., engine failure, battery aging) to DTC sequences. By making an analogy with the natural language processing world and the machine-generated data, the authors could derive when what error pattern is most likely to happen to a vehicle using solely a posteriori error patterns, thus predicting the remaining useful life of a vehicle using DTC sequences. In this paper, we will build upon \cite{math2024harnessingeventsensorydata, Hafeez2024DTCTranGruIT} to further model these multivariate vehicle event sequences and focus on classifying error patterns to benchmark the proposed methods.

\subsection{Event Sequence Modeling}
\noindent Building event sequence models from event data has a plethora of practical applications in several domains. For example, in healthcare, event models ingest electronic health records (EHR) to give diagnoses and treatment plans \cite{pmlr-v219-labach23a, MedBERT}. In cyber security, event models detect intrusion and prevent potential cyber-attacks \cite{MANOCCHIO2024122564}. 

This data is commonly composed of a time of occurrence $t \in \mathbb{R}^+$ and an event type $u \in U$ constructing a pair $(t, u)$ where $U$ is a finite set of discrete event types. A sequence is then constructed with multiple pairs such as $S = \{(t_1, u_1), \ldots, (t_L, u_L)\}$ where $0 < t_1 < \ldots < t_n$. When having more features than the event type and its time of occurrence, we usually refer to it as a \emph{multivariate event sequence}. In our case, we will also have environmental conditions per event type $u$. These discrete sequential data are usually modeled by Hawkes Process \cite{hawkeppp}, where we want to estimate the probability of the next event $(u', t')$ given the history of events $H_t := \{(t_i, u_i) \in \mathbb{R}^+ \times U \mid t_i < t\}$. Neural-based approaches (RNN, LSTM, Transformer) \cite{rrnembedding, selfatthawke} have emerged naturally with the advancement of NLP but come with adaptations proper to event sequences, showing promising results across predictive benchmarks \cite{shou2024selfsupervisedcontrastivepretrainingmultivariate}.

The majority of research utilizing Transformer models for sequence data employs the architecture introduced by \cite{attention}. We define the input embedding $\boldsymbol{U} = \boldsymbol{E} + \boldsymbol{PE} \in \mathbb{R}^{L \times d}$, where $\boldsymbol{E}$ is the event type embedding and $\boldsymbol{PE}$ a positional embedding (usually learnable) indicating the position of each token. We define three linear projection matrices, namely query $\boldsymbol{Q} = \boldsymbol{U} \boldsymbol{W}_Q$, key $\boldsymbol{K} = \boldsymbol{U} \boldsymbol{W}_K$, and value $\boldsymbol{V} = \boldsymbol{U} \boldsymbol{W}_V$, where $\boldsymbol{W}_Q$, $\boldsymbol{W}_K$, and $\boldsymbol{W}_V$ are trainable weights. Essentially, $\boldsymbol{Q}$ represents what the model is looking for based on the input $\boldsymbol{U, K}$, which is the label for the input’s information, and $\boldsymbol{V}$ is the desired
representation of the input $\boldsymbol{U}$.
The attention score $\boldsymbol{A}$ is computed as:
\begin{equation}
\boldsymbol{A} = \text{softmax}\left(\frac{\boldsymbol{Q} \boldsymbol{K}^T}{\sqrt{d}}\right)
\end{equation}

where $d$ denotes the hidden size and $\boldsymbol{A} \in \mathbb{R}^{L \times L}$ the attention scores of each event pair $i, j$. A new representation is computed by

\begin{equation}
\boldsymbol{C} = \boldsymbol{A} \boldsymbol{V}
\end{equation}

A final hidden representation $\boldsymbol{H}$ is obtained via a layer normalization (usually LayerNorm), a pointwise feed-forward neural network (FFN), and residual connections via:
\begin{equation}
\boldsymbol{C}' = \textit{LayerNorm}(\boldsymbol{C} + \boldsymbol{U})
\end{equation}
\begin{equation}
\boldsymbol{H} = \textit{LayerNorm}(\boldsymbol{C}' + \text{FFN}(\boldsymbol{C'}))
\end{equation}
$\boldsymbol{H} \in \mathbb{R}^{L \times d}$ is then used by prediction heads for masked language modeling or next token prediction. On downstream tasks, we usually take the first logit $\boldsymbol{H}_0$ for sequence classification (corresponding to the [CLS] token introduced in BERT \cite{bert}) or the [EOS] (''end of sequence'') for autoregressive Transformers. For the last case, a causal mask is applied to $\boldsymbol{A}$ such that the tokens can only attend to the previous ones.

\subsection{Multimodal Fusion \& Learning}
\noindent Introducing other modalities in a Transformer model has been recently heavily researched \cite{transformermultimodallearningreview, transformerbottleneck}. 
In NLP and more generally in event sequence modeling,
one common and simple way to fuse two token embeddings $\boldsymbol{X}_A \in \mathbb{R}^{L_a \times d}$ and $\boldsymbol{X}_B \in \mathbb{R}^{L_b \times d}$ from modality A and B is via \emph{early summation}. Specifically, \emph{token-wise weighted summing} of multiple embeddings at the input level is defined as: 
\begin{equation}
\boldsymbol{U} = \alpha \boldsymbol{X}_a \oplus \beta \boldsymbol{X}_b.  
\end{equation}
In MedBERT \cite{MedBERT}, the authors summed three types of embeddings (diagnosis codes, the order, and the position of each visit) to form input $\boldsymbol{U}$. This method has several advantages: it is simple to integrate into a Transformer and does not significantly alter the computation. However, the different tokens need to be aligned or projected to a latent space if $L_a \neq L_b$, and it is not clear how one should weigh the different embeddings when summing using $\alpha, \beta$. For example, BERT uses an early summation of its token embedding $\mathbf{E}$ and its position $\mathbf{PE}$ such that $\boldsymbol{U} = \boldsymbol{E} + \boldsymbol{PE}$.

In our case, DTC-TranGRU \cite{Hafeez2024DTCTranGruIT} uses for the three DTC components three separate embeddings that they concatenate along $d$ to form a global DTC embedding $\boldsymbol{D} \in \mathrm{R}^{L \times d}$ (we refer to this method as \emph{early concatenation}).

\begin{equation}
    \boldsymbol{U} = \textit{concat}(\boldsymbol{X}_a, \boldsymbol{X}_b.) \in \mathbb{R}^{L \times (d_b+d_a)}
\end{equation}
\noindent This enables a more fine-grained integration of additional features by selecting their embedding sizes $d_a, d_b$. It also preserves the distinct characteristics of each feature by maintaining separate dimensions, thus potentially learning more nuanced representations and interactions in the deeper attention layers. However, it usually requires more computation since we concatenate along $d$, resulting in more parameters \cite{transformermultimodallearningreview} for the point-wise feed-forward layers and the projection matrices. It also doesn't work well with low cardinality features such as token type ids from BERT \cite{bert} since it will only span across a very small size in the hidden dimension $d$, thus early summation is preferable for this case. By combining embeddings at an early stage, models can learn a unified representation that captures the interactions between different features. This can be particularly useful if the features are highly correlated or if their interactions are crucial for making accurate predictions. These methods are used in a wide variety of domains, in particular in medicine \cite{MedBERT, multidimpatientacuityestimation} with electronic health records (EHR).
When dealing with complex modalities such as audio or images that are not necessarily aligned, the attention mechanism of Transformers \cite{attention} is usually used for fusion. Late fusion in Transformers usually involves encoding the different modalities through independent Transformer encoders to extract high-level representations $\boldsymbol{Z}_a, \boldsymbol{Z}_b$:
\begin{align*}
     \boldsymbol{Z_a} &= \textit{Tf}_a(X_a) \in \mathbb{R}^{L_a\times d}
     \\
    \boldsymbol{Z}_b &= \textit{Tf}_b(X_b) \in \mathbb{R}^{L_b \times d}
\end{align*}
Then, these representations can be concatenated and encoded via another third Transformer (\emph{Hierarchical Attention}) to output a fused representation using multi-head attention:
\begin{equation*}
\boldsymbol{Z} = \textit{Tf}(\textit{concat}(\boldsymbol{Z}_a, \boldsymbol{Z}_b)) \in \mathbb{R}^{L_a \times (d_a + d_b)}
\end{equation*}

\noindent One can also fuse $\boldsymbol{Z}_a, \boldsymbol{Z}_b$ with similarity products like in CLIP \cite{clip}.
The CLIP model introduces a new self-supervised multimodal learning task where the model learns which caption goes with which image and demonstrates SOTA performance for 30 different existing computer vision datasets. The problem with this method is its limited cross-modal interaction, where we first encode separately the modalities to fuse them later. Maybe the model would benefit to already learn the multimodal dependencies earlier in the Transformer. 

On the other end, middle fusion (or mid fusion) methods involve \emph{cross-attention} and more generally \emph{co-attention} mechanisms. In ViLBERT \cite{crossattention} and LXMBERT \cite{lxmbert}, the attention models enable computation of the attention scores $\boldsymbol{A}$ as a function of the image and text input. For example, using $\boldsymbol{X}_a$ as query $\boldsymbol{Q}_a$ and $\boldsymbol{X}_b$ as $\boldsymbol{K}_b, \boldsymbol{V}_b$, we can compute the resulting $\boldsymbol{C}_{a \longrightarrow b} \in \mathbb{R}^{L_a \times d}$ and vice versa (one input attends to another). In \emph{co-attention} both inputs attend to each other by computing two attentions simultaneously ($\boldsymbol{C}_{a \longrightarrow b}, \boldsymbol{C}_{b \longrightarrow a}$), which enables multimodal learning in both ways. One caveat to these methods is that by not projecting $\boldsymbol{X}_a, \boldsymbol{X}_b$ into a fixed latent space $d$ but computing a new $\boldsymbol{A} \in \mathbb{R}^{L_a \times L_b}$, if $L_b \gg L_a$, we drastically increase the computational overhead due to the quadratic time complexity $O(n^2)$ of the \emph{vanilla} attention. We also get two hidden representations for each modality: $\boldsymbol{H}_{a}, \boldsymbol{H}_{b}$. The product $\boldsymbol{Q}.\boldsymbol{K}^T$ carries most of the computation, thus one may consider selecting specific query and key multiplications, resulting in a sparse attention variant like LongFormer \cite{beltagy2020longformer}, or BigBird \cite{bigbird}. These models use a sliding, global, and random attention combination, reducing $O(n^2)$ to a linear complexity of $O(w \times n)$ for LongFormer. Another method is to project the keys and values into a lower-dimensional space, like in Linformer \cite{wang2020linformer} or use a multimodal bottleneck transformer \cite{transformerbottleneck} to compute the cross-attention in a restricted latent space.

We note that fusing features in event sequence models is not well-researched. The majority of papers aim at fusing the time $t_i$ information per event type $u_i$ by performing an \emph{early summation} \cite{Zhou2023, selfatthawke, transformerhawkeprocess} to integrate the time component, sometimes omitting the positional embedding for the time embedding \cite{transformerhawkeprocess}, sometimes summing both \cite{shou2024selfsupervisedcontrastivepretrainingmultivariate, sparsetemporalattention}, or not taking it into account \cite{Hafeez2024DTCTranGruIT}.
\section{Data}
\subsection*{DTC: Diagnosis Trouble Code}
\noindent Diagnostic data is generated by various Electronic Control Units (ECU) in a vehicle at irregular intervals or concurrently. Diagnostic data differs substantially from raw sensor data since diagnostic data is categorical and relates to various problems and software statuses within the vehicle. A Diagnosis Trouble Code (DTC) is constructed from three pieces of information occurring at the same timestamp $t_s$ and mileage $m_a$, namely (1) the ID number of the ECU, (2) an error code (Base-DTC), and (3) a Fault-Byte.
A single DTC is composed of these three elements such that:
\[
\text{DTC} = \text{ECU} \mid \text{Base-DTC} \mid \text{Fault-Byte}.
\]
This research uses an anonymized vehicular DTC sequence dataset of 5 million sequences with on average $L \approx 150 \pm 90$ DTCs per sequence. Each sequence $S_{raw}$ belongs to a unique vehicle. In $S_{raw}$, each DTC (commonly referred to as event type $u_i$) is attached with
\begin{itemize}
    \item spatiotemporal features time $t_i$ and mileage $m_i$. 
    \item a small sequence $S^e_i = \{(d_j, v_j, c_j)\}_{j=0}^{L^e_i}$ of environmental conditions, respectively with elements description, value, and unit. 
\end{itemize}
These elements construct a single event $(u_i, t_i, m_i, S^e_i)$. To obtain a full sequence $S_{raw} = \{(u_i, t_i, m_i, S^e_i)\}_{i=0}^L$, we obtain the last known timestamp $t_{sL}$ and mileage $m_{aL}$ and select all DTCs that are no further than (1) a given period in the past ($t_{sL} - t_{si} < 30$ days), and (2) a given distance in the past ($m_{aL} - m_{ai} \leq 300$ km).
We then split $S_{raw}$ into two distinct sequences: $S = \{(u_i, t_i, m_i)\}_{i=0}^L$ for the DTCs, and $S^e = \{(d_i, v_i, c_i)\}^{L^e}_{i=0}$ for only the environmental conditions with length $L^e \approx 2275 \pm 2310$. An overview of the elements is provided in Table~\ref{notation} as well as the number of distinct values for each feature in Table~\ref{distinct_value_des}. 
\begin{table}[ht!]
\begin{center}
\caption{List of symbols and their respective descriptions.}
\label{notation}
    \begin{tabular}{ll}
        \hline
        \textbf{Notations} & \textbf{Description} \\
        \hline
        $u$ & Discrete event type \\
        $m_a$ & Absolute mileage of the vehicle (km) \\
        $t_s$ & Unix timestamp attached to each DTC \\
        $S$ & Sequence of triplets from DTC codes defined \\
            & as $S = \{(u_i, t_i, m_i)\}_{i=0}^L$ of length $L$ \\
        $S^e$ & Sequence of environmental conditions triplets defined as \\
             & $S^e = \{(d_i, v_i, c_i)\}_{i=0}^{L^e}$ of length $L^e \gg L$ \\
        $d$ & Environmental condition description, e.g., temperature \\     
            & increase, vehicle speed, pressure increase \\
        $v$ & Environmental condition value can be int, float, string, null \\
        $c$ & Environmental condition unit, e.g., A, V, bar, °C, sec\\
        \hline
    \end{tabular}
    \end{center}
\end{table}
\subsection{Environmental Conditions} \label{data_env}
\noindent Integrating sensory information efficiently, such as temperature or pressure values for Predictive Maintenance (PdM), alongside discrete codes remains an open problem \cite{Zhou2023, Hafeez2024DTCTranGruIT}. Intuitively, it might seem trivial to rely on temperature for diagnosing a specific defect in a vehicle engine or a voltage measurement in a battery to identify a cell failure. Domain experts often analyze this data to make decisions about EPs in vehicles. However, the integration of environmental conditions comes with three significant challenges:
\begin{enumerate}
    \item \emph{Dimensionality:} descriptions have a high cardinality ($>10^3)$ and values can be multi-types: strings, booleans, integers, floats, or NaN.
    \item \emph{Variability:} differs across each DTC: e.g., we cannot be certain whether we will observe 'temperature' for DTC1 and 'pressure' for DTC2.
    \item \emph{Volume:} there are multiple e. conditions per DTC, which can be redundant, noisy, and duplicated across the sequence $S^e$.
\end{enumerate}
The \emph{Dimensionality} problem increases model complexity and often necessitates heavy data engineering techniques beforehand. The \emph{Variability} makes integration into a machine learning model challenging, and lastly, the \emph{Volume} requires substantial data infrastructure to manage e. conditions, since it is often several times the size of DTC data and significantly increases the sequence length, a known challenge in Transformers \cite{sparsetemporalattention}. We partially address these challenges by:
\begin{itemize}
    \item removing \textit{null} values, units, or descriptions.
    \item removing duplicates per DTC: the same $(d, v, c)$ triplet occurring multiple times within the same $S^e_i$ is removed, retaining only one instance. This significantly reduced redundancy and noise in the data, though we still observed $L^e \approx 2275 \pm 2310$ triplets per $S^e$. One caveat of this approach is the \textit{nonalignment} of e. conditions and DTC when multiple DTCs occur at the same $t_i$. In such cases, we drop subsequent e. conditions after the first occurrence.
    \item selecting the top-$18$ most popular units.
\end{itemize}

\begin{table}
\center
\caption{Number of distinct values for each feature.}
\label{distinct_value_des}
\begin{tabular}{|l|r|p{3cm}|}
\hline
\textbf{Data} & \textbf{\# of values} & \textbf{Description} \\ \hline
DTC & 22,137 & Diagnostic Trouble Code \\ 
ECU & 132 & Electronic Control Unit \\ 
Base-DTC & 17,044 & Error Code \\ 
Fault-Byte & 2 & Binary Value \\ 
E. Condition Description & 2,559 & (See Table 1: $d$) \\ 
E. Condition Value & 3,288 & (See Table 1: $v$) \\ 
E. Condition Unit & 18 & Most popular units: $u$) \\ \hline
\end{tabular}
\end{table}

\noindent To ingest the resulting $S^e$ into a Transformer, we map our continuous values $v \in \mathbb{R}^+$ into discrete \textit{tokens} \cite{transformermultimodallearningreview} with a limited vocabulary. Thus, we search for the optimal bins for each unit $c$. A variation of the Greenwald-Khanna algorithm \cite{spaceefficientonlinecomputationofquantile} with memory requirement in the worst case of $O(\frac{1}{\epsilon}\log{(\epsilon N)})$ is used with precision $\epsilon=0.0001$, up to a maximum of $\theta = 4000$ tokens per unit $u$ to limit a potential $18 \times \theta = 72000$ values. 
In practice, they overlap and don't reach the maximum (Table \ref{distinct_value_des}).
\section{BiCarFormer}
\begin{figure}[!b]
      \centering
      \includegraphics[width=3.4in]{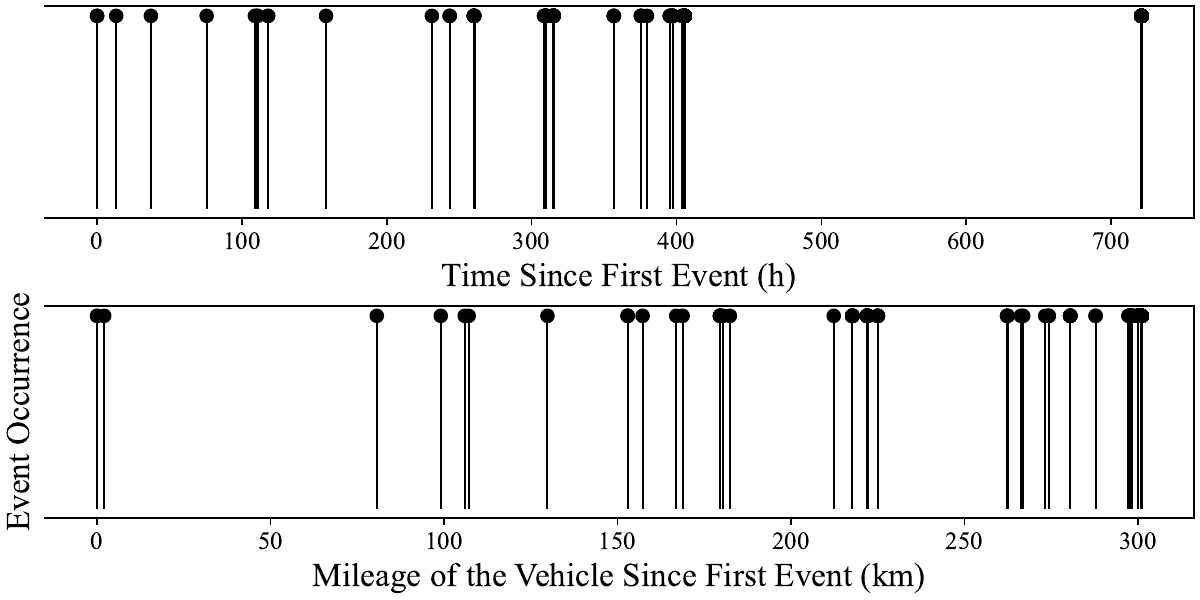}
      \caption{Temporal and spatial point process representation of events from a vehicle. Bold vertical lines indicate multiple events happening at the same time $t_i$ or mileage $m_i$.}
      \label{fig:tpp}
\end{figure}
\begin{figure*}[!t]
    \centering
    \includegraphics[width=6.5in]{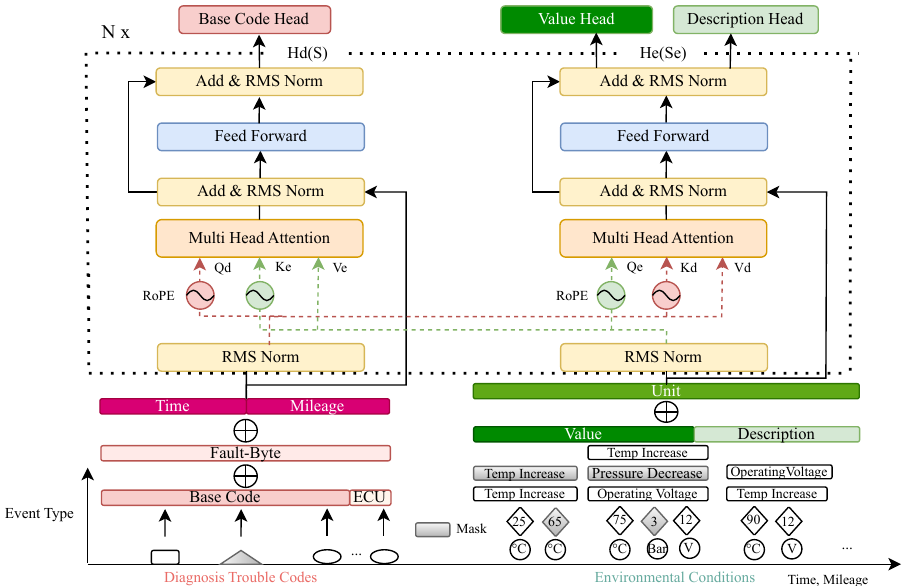}
    \caption{BiCarFormer overall architecture with multimodal masking. Both parallel transformers are computing cross attention scores conditioned on each modality $\boldsymbol{Q, K, V}$. Two final representations are generated for each modality: DTC ($\boldsymbol{H}_d)$ and e. conditions ($\boldsymbol{H}_e$). Multiple embeddings are defined at the input level to take into account token-specific features.}
    \label{bicarformer}
\end{figure*}
\subsection{DTC Embeddings}
\label{sec:dtc_model_section}
\noindent We use a Bidirectional Transformer model \cite{bert} that we train with mask language modeling. Each DTC element is embedded in a specific feature space, namely: \emph{ECU} ($\boldsymbol{D}_{ecu} \in \mathbb{R}^{L \times d_{ecu}}$), \emph{Base-DTC} ($\boldsymbol{D}_{base} \in \mathbb{R}^{L \times d_{base}}$), and the \emph{Fault-Byte} ($\boldsymbol{D}_f \in \mathbb{R}^{L \times d}$) using separate lookup tables. The first two are concatenated along the feature dimension, like in \cite{Hafeez2024DTCTranGruIT}, and $\boldsymbol{D}_f$ is added to the result like a token type id in BERT \cite{bert}. This reduces the vocabulary size and trainable parameters compared to \cite{math2024harnessingeventsensorydata} and ensures that the embeddings with high cardinality are preserved independently, thus capturing more relationships between the \textit{ECU}, the \textit{Base-DTC}, and the \textit{Fault-byte}. We obtain the input DTC embedding $\boldsymbol{D} \in \mathbb{R}^{L \times d}$: 
\[
\boldsymbol{D} = \text{concat}(\boldsymbol{D}_{ecu}, \boldsymbol{D}_{base}) + \boldsymbol{D}_f
\]
During the pretraining phase, we mask the Base-DTC token only, since it serves as primary information.
\section{Positional Embeddings}
\noindent Understanding the positioning of failure events in both time and mileage is beneficial for predictive maintenance. Failure events in our dataset exhibit spatial and temporal patterns (Figure \ref{fig:tpp}), which could indicate stationary behavior of the vehicle or recurring temporal failures \cite{math2024harnessingeventsensorydata}. Prior work, such as \cite{Zhou2023}, employs adaptive binning to discretize time in long event sequences. However, given our moderate sequence length (258 DTCs with 30-day and 300-km intervals), binning is unnecessary. Additionally, lookup table embeddings, often used in discrete event modeling, would be computationally expensive, introduce unnecessary parameters, and fail to preserve meaningful distance relationships in $\mathbb{R}^+$. 

\noindent We adopt a continuous time embedding \cite{selfatthawke}. Specifically, given the absolute time $t_i$ and mileage $m_i$ of event $u_i$, we define the time embedding as:
\begin{equation}
\mathbf{T}_{i,j} := \begin{cases}
    \sin(t_i \times \theta_{0, u}^{j/d}) & \text{if } j\mod 2 = 0 \\
    \cos(t_i \times \theta_{0, u}^{(j-1)/d}) & \text{if } j\mod 2 = 1\\
\end{cases}
\label{eq:time_mileage}
\end{equation}
We reuse Eq \eqref{eq:time_mileage} to create a mileage embedding $\boldsymbol{M} \in \mathbb{R}^{L \times d/2}$.
Both embeddings, $\boldsymbol{T}, \boldsymbol{M}$, are concatenated along the feature dimension to form a unified positional representation. This preserves their independent contributions and prevents interference when fused with $\boldsymbol{D}$. Our final fused input $\boldsymbol{U} \in \mathbb{R}^{L \times d}$ is given by: 
\begin{equation}
\boldsymbol{U} = \text{concat}(\boldsymbol{D}_{base}, \boldsymbol{D}_{ecu}) + \boldsymbol{D}_f +  \text{concat}(\boldsymbol{T, M})
\end{equation} 
\subsection{Environmental Embeddings}
\noindent To better capture relationships between the different e. conditions elements $(d, v, u)$, we create three distinct learnable embeddings: (1) the description of the e. conditions $\boldsymbol{D}_e \in \mathbb{R}^{L^e \times d_d}$ (2) the discretized value $\boldsymbol{V}_e \in \mathbb{R}^{L^e \times d_v}$ (3) its unit $\boldsymbol{U}_e \in \mathbb{R}^{L^e \times d}$.
We use a mix of early summation and concatenation to fuse these embeddings at the input level and obtain the total e. conditions embedding $\boldsymbol{E} \in \mathbb{R}^{L^e \times d}$:
\[
\boldsymbol{E} = \text{concat}(\boldsymbol{V}_e + \boldsymbol{D}_e) + \boldsymbol{U}_e
\]
To fuse the e. conditions, we cannot simply align them with their respective DTC to perform an early token-wise summation or concatenation with $\boldsymbol{D}$. Since we are facing real-world data (noisy, redundant) and doing heavy data filtering, there are too many missing e. conditions for each DTC.
Thus, we chose to create a separate sequence $S_e$ that is much longer than $S$, where we will concatenate all e. conditions and fuse them into one embedding $\boldsymbol{E}$. This straightforward method is flexible and $\boldsymbol{E}$ can be employed in a middle fusion manner inside the attention mechanism \cite{transformermultimodallearningreview}.

\subsection{Co-attention for Vehicle Event Sequences}
\noindent The overall architecture of BiCarFormer is shown in Figure \ref{bicarformer}. 
Our architecture is directly inspired by the co-attention mechanism of ViLBERT \cite{vilbert}, which enables multimodal learning by computing attention scores conditioned on each modality. 
Two multi-head attention layers are processing $S$ and $S^e$, resulting in two attention scores computation conditioned on: $\boldsymbol{D}$ (DTCs) and $ \boldsymbol{E}$ (e. conditions) fused embeddings. 
We apply a RoPE (Rotary Position Encoding) \cite{Roformer} on two sets of queries and keys (one for each cross attention) to induce the absolute and relative position of tokens (Figure \ref{bicarformer}). 

More specifically, the fused input embedding vectors $\boldsymbol{u}_m \in \mathbb{R}^d$ and $\boldsymbol{e}_n \in \mathbb{R}^d$ from token positions $m \in [1, L], n \in [1, L^e]$ are projected through weights $\boldsymbol{W}_q, \boldsymbol{W}_k, \boldsymbol{W}_v$.
Then, in complex coordinates, the query and key are given by:
\begin{align}
\boldsymbol{q}_m &= e^{im\theta_u} \boldsymbol{W}_q \boldsymbol{u}_m \\    
\boldsymbol{k}_n &= e^{in\theta_e} \boldsymbol{W}_k \boldsymbol{e}_n
\end{align}
where $\theta_u = \text{diag}(\theta_1, \dots, \theta_{d/2})$ with $\theta_{i} = \theta_{0,u}^{-2i/d}$, $\theta_{0, u} = 5000$. Same for $\theta_e$ but with $\theta_{0, e} = 80000$ due to $L^e \gg L$. We use an alignment function \cite{survey_att_mechanisms} $f: \mathbb{R} \rightarrow [0, 1]$ to produce the attention weights between two tokens:
\begin{align}
    a(\boldsymbol{u}_m, \boldsymbol{e}_n, m,n) &= f(\boldsymbol{q}_m^T \boldsymbol{k}_n) = a_{dtc \rightarrow env}\\
    a(\boldsymbol{u}_m, \boldsymbol{e}_n, m, n) &= f\left(\boldsymbol{u}_m^T \boldsymbol{W}_q^T e^{i(n\theta_e - m\theta_u)}\boldsymbol{W}_k \boldsymbol{e}_n\right) \label{eq:rope_q_k}
\end{align}
We also compute Eq \eqref{eq:rope_q_k} for $a_{env \rightarrow dtc} = a(\boldsymbol{e}_m, \boldsymbol{u}_n, n, m)$ with separate weights for $\boldsymbol{q}_m, \boldsymbol{k}_n$. Finally, two cross-attended context vectors are produced:
\begin{align}
    \boldsymbol{c}_{dtc \rightarrow env}(m) &= \sum^{L^e}_{n=1} a_{dtc \rightarrow env}(m,n) \boldsymbol{v}_{e,n} \label{eq:coatt_dtc_env} \in \mathbb{R}^d\\
    \boldsymbol{c}_{env \rightarrow dtc}(n) &= \sum^{L}_{m=1} a_{env \rightarrow dtc}(n,m) \boldsymbol{v}_{u, m} \label{eq:coatt_env_dtc} \in \mathbb{R}^d
\end{align}
where values $\boldsymbol{v}_u, \boldsymbol{v}_e$ are obtained from $\boldsymbol{V}_e = \boldsymbol{W}^e_{v} \boldsymbol{E}, \boldsymbol{V}_u = \boldsymbol{W}^u_v \boldsymbol{D}$.
\noindent We tried two different alignment functions $f$ \cite{survey_att_mechanisms} for Eq \eqref{eq:rope_q_k}. The widely used \textit{softmax} and \textit{1.5-entmax} \cite{peters-etal-2019-sparse}.
The standard \emph{softmax} outputs dense attention scores which take too many e. conditions into account. Since our input data is noisy and redundant, we would like to extract salient information, thus a sparse alignment function seems like an intuitive choice. 
However, in practice, we didn't find a benefit using 1.5-entmax in the pretraining and classification where we observed a 30\% slow down in iterations per second. Thus, we stick with \textit{softmax}.
As a consequence, two hidden states $\boldsymbol{H}_{d} \in \mathbb{R}^{L\times d}$ and $\boldsymbol{H}_{e} \in \mathbb{R}^{L^e\times d}$ (Figure \ref{bicarformer}) for each modality are computed after fully connected layers, residual connections, and root-mean-squared normalizations \cite{rmsnorm, touvron2023llamaopenefficientfoundation}. 
\subsection{Multimodal Learning}
\begin{figure}[!t]
      \centering
      \includegraphics[width=3in]{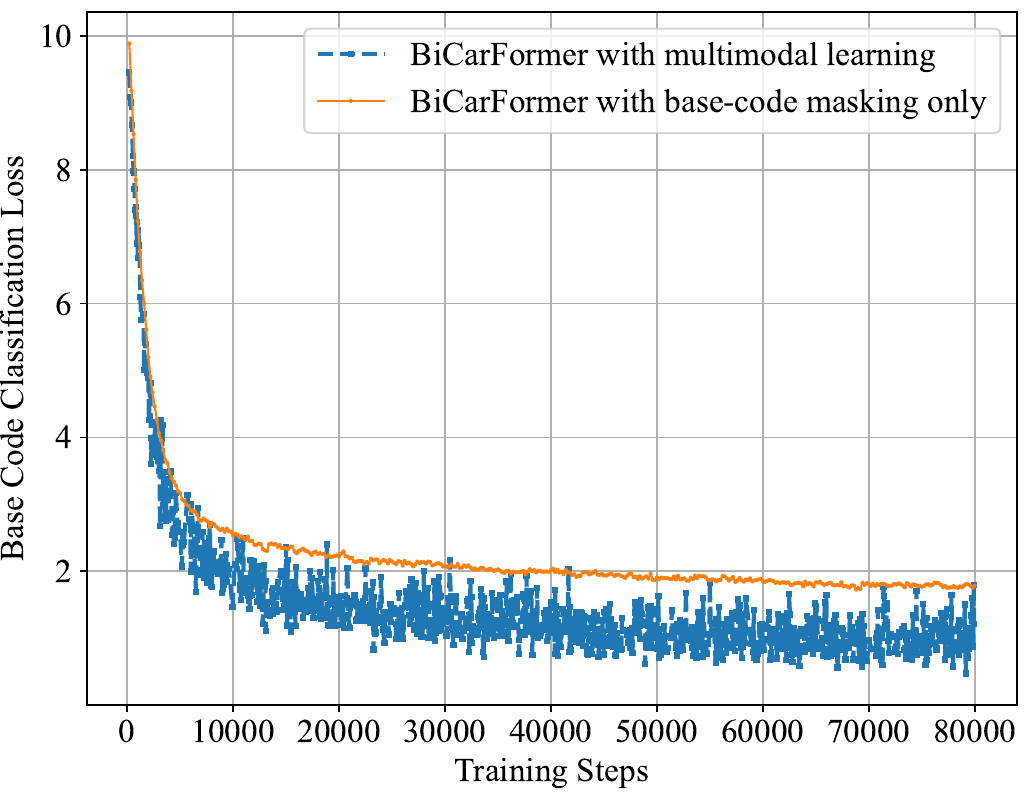}
      \caption{Pretraining Base-DTC classification loss comparison with and without multimodal learning.}
      \label{fig:pretraining_comp}
\end{figure}

\noindent To reinforce the relationship between DTC events and e. conditions, we not only mask the DTC but also the e. conditions duet $(d, v)$ and let the unit unmask to reduce training complexity. This enables multimodal learning by reconstructing $S^e$ using $S$ and vice versa. Hence, BiCarFormer learns to benefit from this extra modality, which can be confirmed by taking a look at the cross entropy loss of the \textit{Base-DTC} classification: $\mathcal{L}_{\text{dtc}}$ where using both modalities helps to reduce the pretraining loss (Figure \ref{fig:pretraining_comp}). However, due to multi-task learning, the losses are less steady and might require gradient clipping and a smaller learning rate to stabilize training. BiCarFormer is trained with three cross-entropy losses balanced by static coefficients: $ \alpha=0.5, \beta=0.3, \gamma=0.2$:
\begin{equation}
    \mathcal{L} = \alpha \mathcal{L}_{\text{dtc}} +  \beta \mathcal{L}_{\text{value}} +  \gamma \mathcal{L}_{\text{description}}
\end{equation}

\section{Experiments}
\noindent \textbf{Settings.}
\noindent 
We evaluated our model against established ''sequence-to-sequence'' Transformer architectures, including BERT \cite{bert} (without the Next Sentence Prediction task) and DTC-TranGRU \cite{Hafeez2024DTCTranGruIT}. Since DTC-TranGRU is autoregressive \cite{Hafeez2024DTCTranGruIT}, we trained it using next token prediction and three separate heads, which classify the three DTC components. To assess each model's performance, we focused on multi-label classification of error patterns on downstream.
For BERT and BiCarFormer we masked 15\% of the tokens during the pretraining. The different models comprised approximately 25 million parameters with the same hidden size $d=600$, which ensures fair comparison.
For our downstream task of multi-label sequence classification, we used either the [CLS] token to perform classification or in the case of the BiCarFormer, the $[CLS]_{dtc}$ and $[CLS]_{env}$ after simply concatenating them along the feature dimension $d$ and fusing with a small $MLP$. We took the [EOS] token for DTC-TranGRU instead of the [CLS].
Every classifier has 3,461,160 trainable parameters and is composed of a small $MLP$ with layer normalization, residual connections, and a sigmoid activation. We froze all backbones during classification. The dataset was partitioned into training, validation, and testing sets, adhering to a ratio of 75\%, 15\%, and 15\%, respectively, with about 360 labeled error patterns. 
For the pretraining, we used a learning rate of $10^{-4}$ with a cosine scheduler and warm-up of 2000 steps for all models. We used the AdamW optimizer \cite{loshchilov2018decoupled} with $\beta_1$ set to 0.9 and $\beta_2$ to 0.99. We applied a weight decay of 0.1, and a percentile gradient clipping of 5. Finally, we trained on an NVIDIA A10G GPU using fp16 with a batch size of 32 for the pretraining and 192 for the classifiers. 

\textbf{Metrics}. The evaluations were conducted across 360 different error patterns. To measure classification performance, we employed multiple classification metrics namely the AUROC (Area Under the Receiver Operating Characteristic), Precision, Recall and F1 Score metrics \cite{classificationmetrics}. The last 3 were computed using a confidence threshold of 0.8. The metrics were aggregated using different averaging: micro, macro, and sample.
Micro computes metrics globally and favors frequent classes. It gives a good feeling of how the model performs on the overall data distribution but it does not reflect the performance \textit{per-class} or \textit{per-instance}. In contrast, macro averaging treats all classes equally by computing metrics independently for each class and then averaging the results. However, in multi-label classification, the focus is often on \textit{per-instance} performance rather than \textit{per-class} performance. This means we prioritize the overall quality of multi-label predictions for each instance, rather than optimizing for the accuracy of individual labels in isolation. For this, the sample averaging computes metrics \textit{per-instance} and then averages across all samples.
\begin{table*}[!t]
\begin{center}
\caption{Downstream evaluation of multi-label error pattern classification. Each sequence2sequence model has the same number of parameters (25m). We use different averaging for a more insightful evaluation.}
\label{table_comparaison_auroc}
\begin{tabular}{|c|c|c|c|c|c|c|}
\hline
Model & AUROC (Micro) & F1 Score (Micro) & F1 Score (Macro) & Precision (Sample) & Recall (Sample) & F1 Score (Sample) \\
\hline
\textbf{BiCarFormer} & \textbf{0.809} & \textbf{0.77} & \textbf{0.71} & \textbf{0.68} & \textbf{0.62} & \textbf{0.64} \\
\hline
DTC-TranGRU \cite{Hafeez2024DTCTranGruIT} & 0.602 & 0.36 & 0.28 & 0.23 & 0.19 & 0.2 \\
\hline
BERT \cite{bert} & 0.768 &  0.71 & 0.63 & 0.59 & 0.53 & 0.55 \\
\hline
\end{tabular}
\end{center}
\end{table*}
\noindent  
\subsection{Multi-label Classification Performance Analysis}
\noindent Table \ref{table_comparaison_auroc} presents a comparative analysis of BiCarFormer against established sequence-to-sequence models for multi-label error pattern classification. Multi-label classification is particularly challenging due to the varying complexity of error patterns, where a single instance may be associated with multiple diagnostic trouble codes (DTCs) and environmental conditions. DTC-TranGRU \cite{Hafeez2024DTCTranGruIT} underperforms, particularly in \textit{per-class} and \textit{per-instance} evaluations, likely due to the limitations of masked attention mechanisms and model architecture. These constraints make it difficult for the model to capture long-range dependencies in the input sequence, leading to a lower AUROC (0.602) and poor F1 scores (Micro: 0.36, Macro: 0.28, Sample: 0.2). While BERT \cite{bert} is already achieving relatively high AUROC and F1 Micro, 
BiCarFormer significantly enhances the AUROC (Micro) by +4\%, and the F1 Score (Micro) by +6\%.

\noindent The improvement of BiCarFormer is especially visible for \textit{per-class} averaging. When dealing with many error patterns, some are more difficult to distinguish due to their natural complexity and overlaps. Thus, this might confuse a classifier. 
BiCarFormer, however, better differentiates rare classes, suggesting an improved capacity to generalize across a diverse range of error patterns.
This is confirmed in this experiment, where we see the gaps in classification performance when averaging across classes, with a +8\% F1 Score compared to BERT. This means that classes with small instances, usually hard to classify using only the DTCs, can be better differentiated using e. conditions.
Finally, \textit{per-instance} precision and recall are the biggest improvement with +9\% in F1 Score, Precision, and Recall compared to BERT. These results confirm that BiCarFormer effectively leverages this additional information and highlights the predictive improvement compared to standard ''sequence-to-sequence'' models.
\begin{figure}[h]
      \centering
      \includegraphics[width=3.3in]{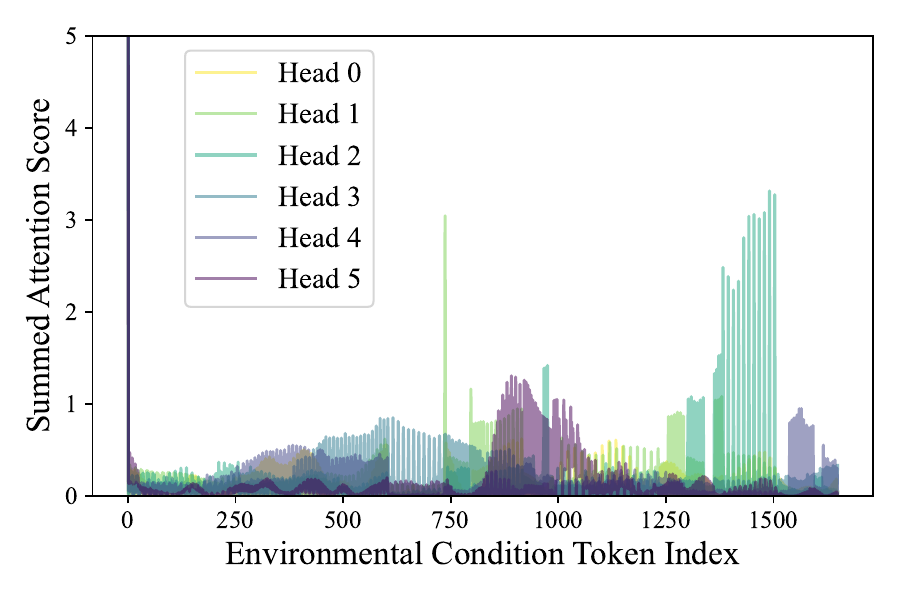}
      \caption{Amount of attention received by each e. condition from the DTCs. The y-axis was truncated to improve clarity as well as the number of heads printed. We take $\boldsymbol{A}_{dtc \rightarrow env}$ of the last layer.}
      \label{fig:att_distrib}
\end{figure}

\begin{figure*}[!t]
\centering    
\includegraphics[width=7.1in]{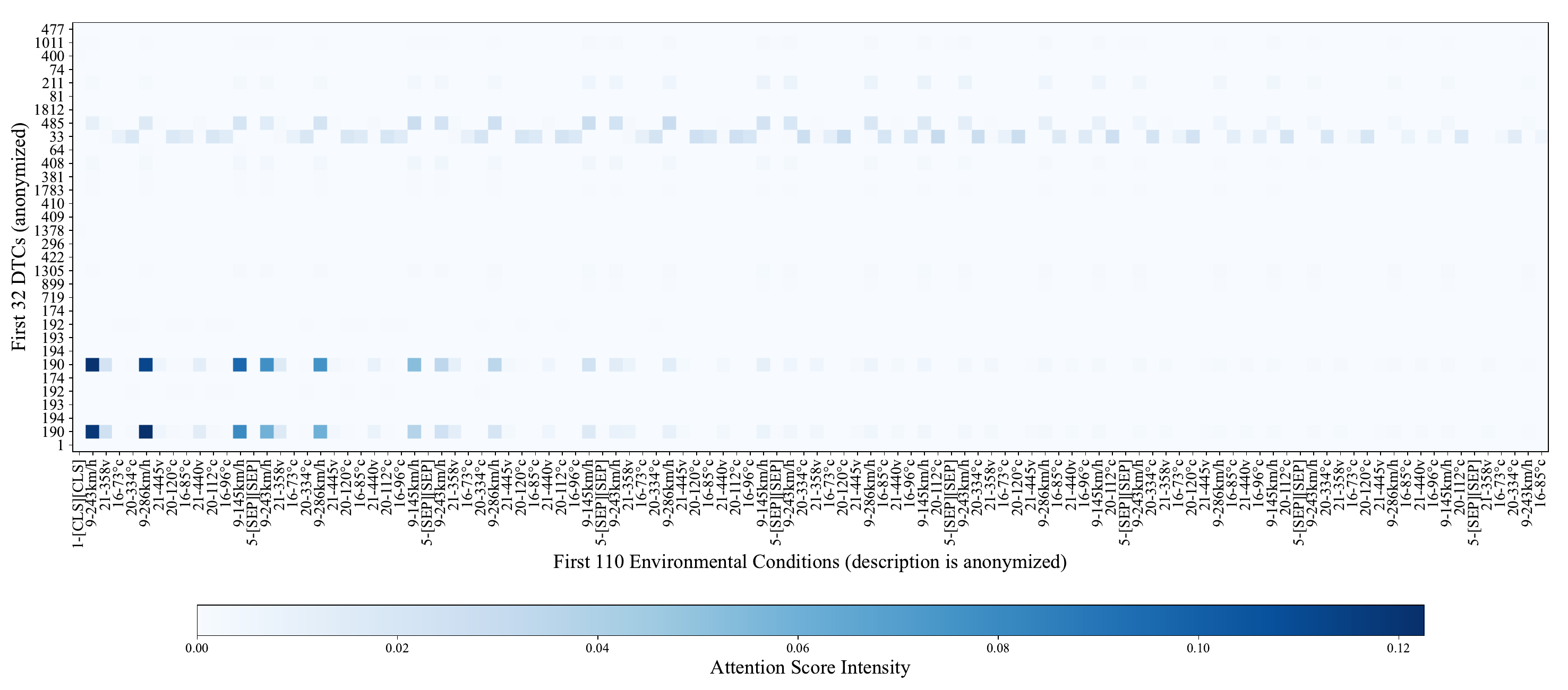}
    \caption{Cross Attention Scores for $\boldsymbol{A}_{dtc \rightarrow env}$. The DTCs are shown on the y-axis (anonymized), and the environmental conditions with their 3 elements $(d, v, u)$ concatenated are shown on the x-axis (the description $d$ is anonymized). The intensity of each cell reflects the attention weight, where darker shades indicate higher attention values.}
    \label{att_scores}
\end{figure*}

\subsection{Cross-Attention Scores Interpretations}
\noindent We would like to understand how the co-attention mechanism enables multimodal learning and enhances downstream tasks. We took a random test sample with a specific battery issue and ran it on BiCarFormer to analyze the different cross-attention score patterns. 
Figure \ref{att_scores} presents a heat-map visualization of $\boldsymbol{A}_{dtc \rightarrow env}$. This helps us to understand general attention patterns from DTCs to e. conditions.
Cross-attention uncovers natural relationships from certain DTCs linked to specific e. conditions. Let's take DTC 190, we can see that this token focuses more on a local series of e. conditions with unit \textit{km/h} rather than other units. Whereas DTC 33 and 485 attend more to \textit{°C}.
This makes sense and is due to specific e. conditions that characterize certain DTCs, i.e., the temperature might define DTC1 while DTC2 might indicate a rise in the voltage.
Due to the duplicated e. conditions unit across $S^e$, one DTC might attend to the next e. conditions 'voltage' also later in the sequence, creating these line patterns on Figure \ref{att_scores} with DTC 190 \& 33 \& 485.
As a consequence, we notice the local temporal patterns between DTCs and their surrounding e. conditions (Figure \ref{att_scores}). Therefore, the initial DTCs should prioritize the initial e. conditions in $Se$, with an offset resulting from the difference in sequence length between $L$ and $Le$. For instance, DTC 190 at positions 1 and 6 should prioritize the initial e. conditions.

We would like to assess if the DTCs attend to a few key e. conditions. From the same test sample, Figure \ref{fig:att_distrib} plots the amount of attention that DTCs give to each of the e. conditions. Notably, we observed different clusters per head. Each head learns to focus on specific e. conditions, sometimes specific indexes like \emph{head 1} ($i \approx 749$ and $800$), and \emph{head 2} ($i \approx 960)$. The [CLS] token is traditionally heavily attended. Interestingly, the last part of the sequence seems to be more attended by specific heads: $i \in [1300, 1500]$, signaling that we can capture long-range dependencies between DTCs and tokens of e. conditions even with a big difference in sequence length (i.e., with $L^e \gg L$). Moreover, late e. conditions might carry a lot of defect signals, which makes sense since we are getting closer to the critical failures happening at time $t_i = t_L$. Thus a rise in the attention scores is observed.  
\begin{figure}[!b]
      \centering
      \includegraphics[width=3.2in]{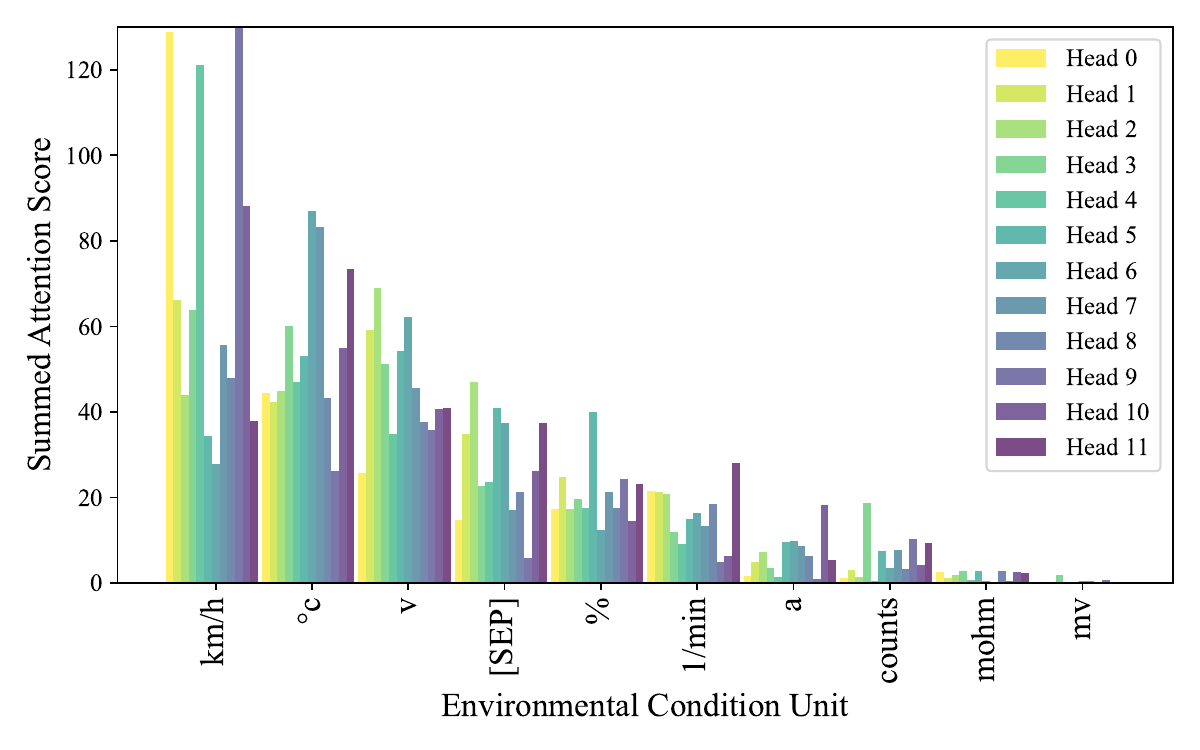}
      \caption{Amount of attention received by each environmental unit from the DTCs in a \textit{battery aging} error pattern from a battery electrical vehicle.  We take $\boldsymbol{A}_{dtc \rightarrow env}$ of the last layer and print all heads.}
      \label{fig:att_distrib_value}
\end{figure}
\subsection{Error Pattern in Battery Electrical Vehicles}

\begin{figure}[h]
      \centering
      \includegraphics[width=3.3in]{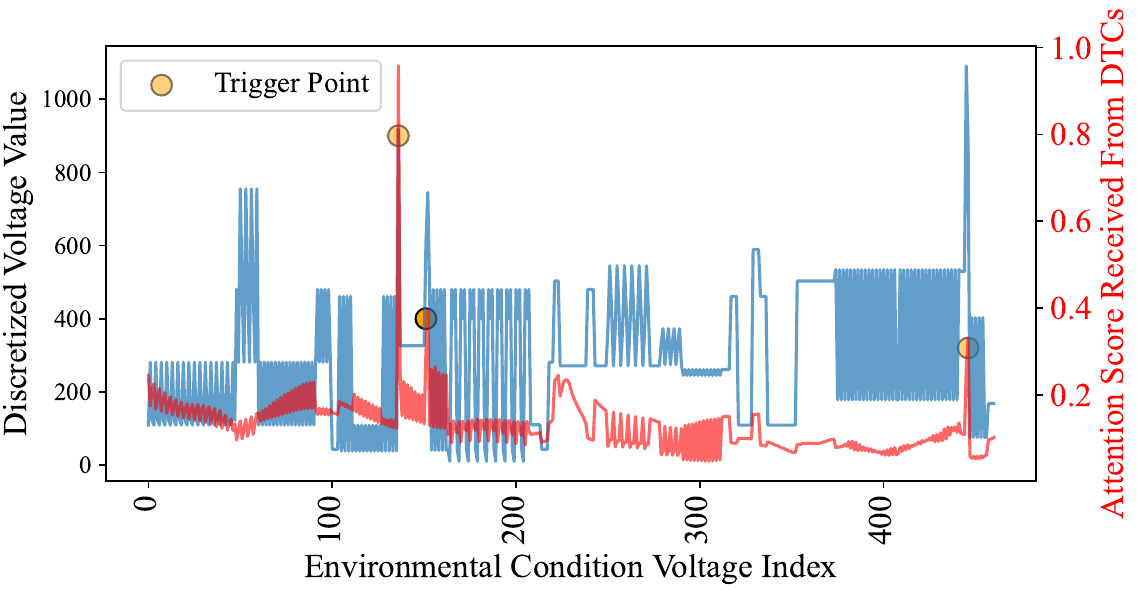}
      \caption{Discretized voltage variation in function of environmental condition tokens and the associated amount of attention received from DTCs. We only take triplets $(d,v, u)$ from $S^e$ with $u = $ 'v' and extract $\boldsymbol{A}_{dtc \rightarrow env}$ from the last layer and \emph{head 2}. The sample is taken from a \emph{battery aging} error pattern.}
      \label{fig:att_distrib_unit}
\end{figure}

\noindent With the increasing popularity of \textit{BEVs} (battery electrical vehicles), there exists an extensive need to diagnose failures within these vehicle types.
We take the example of a concrete error pattern such as a \textit{battery aging}, which is a common phenomenon in electrical vehicles. And draw the cumulative attention of e. condition units for the same test sample as before.
Since attention heads focus on a certain part of $S^e$ (Figure \ref{fig:att_distrib}), some should focus on special units (Figure \ref{fig:att_distrib_unit}) such as km/h (\emph{head 0, 4, 9}) while other on °C (\emph{head 6, 7}) and voltage (\emph{head 1, 2}).
Now, more interestingly, we take a head that tends to specialize in one unit like \emph{head 2} for voltage and draw the evolution of the discretized voltage value in function of their index in $S^e$ in Figure \ref{fig:att_distrib_value}. 
Thus, we essentially try to answer if we have a fluctuation of voltage values and a correlated change in the attention pattern.
We also draw the associated attention scores received from DTCs and obtain multiple \emph{trigger points} where the voltage value strongly correlates with attention scores. We note that this is not the same depending on the heads, some exhibit exponentially decreasing attention scores shapes while other noisy signals. 
As a consequence, BiCarFormer is able to extract specific fluctuations of continuous voltage values from a tokenized input. Hence, one may extract trigger points and analyze how these specific variations affect the vehicle, providing a more explainable result to domain experts.

\section{Applicative Downstream Tasks}
\noindent By having a separate hidden state for the e. conditions, one could also apply it for unsupervised learning where we don't know some EPs which is a common phenomenon. Thus, instead of having just $H_d$, one could rely on $H_e$ to provide a more accurate dimensionality reduction and observe better decision boundaries between unlabeled EPs using clustering techniques. Moreover, if provided with domain knowledge, we could directly filter $S^e$ beforehand to take specific e. conditions rather than injecting a significant amount. Our model and overall paradigm make it possible to keep this architecture and reduce the computation by injecting domain knowledge directly into the preprocessing steps, making it easy to compare the results. 
Another possible useful downstream application is explainability. Where we would like to see the contributions of e. conditions and DTCs to the EPs classification. Since error patterns are hard-coded defined rules by domain experts, we could try to derive new rules for unknown error patterns based on feature attribution methods \cite{shap}, cross-attention scores, perturbation-based methods or multi-label causal discovery with Transformers \cite{alonso2024transformers}.
Contrary to \cite{math2024harnessingeventsensorydata}, BiCarFormer model classifies EPs a posteriori, thus the inference optimization might not be needed since we don't rely on edge computing capabilities within the vehicles. Nevertheless, due to the quadratic time complexity of cross-attention scores, there is an extensive need to optimize Eq \eqref{eq:coatt_dtc_env}, and \eqref{eq:coatt_env_dtc}. Alternatives like sparse attention \cite{sparsetemporalattention, bigbird, beltagy2020longformer}, token merging \cite{bolya2023token} or pooling should be adapted for co-attention in event sequences.
\section{Conclusion}
\noindent 
In this work, we introduced BiCarFormer, a novel multimodal transformer-based model designed to capture the intricate relationships between Diagnostic Trouble Codes (DTCs) and environmental conditions in modern vehicles. Through cross-attention mechanisms, we demonstrated how BiCarFormer effectively learns cross-modal interaction between error codes and their environmental conditions, leading to improved classification performance and more interpretable attention patterns. By interpreting attention scores, domain experts can understand BiCarFormer predictions and verify or derive new rules for error patterns. 
The model’s superior performance in \textit{per-class} and \textit{per-instance} evaluations highlight its potential for real-world deployment in diagnostic applications where accurate multi-label predictions are essential.
Beyond classification, BiCarFormer offers promising directions for unsupervised anomaly detection of error patterns, explainability, and domain-knowledge-oriented models.
Future work may explore more efficient co-attention to handle long environmental condition sequences.
\IEEEpubidadjcol
\bibliographystyle{IEEEtran}
\bibliography{math}

\vfill
\end{document}